**Federated Causal Inference in Healthcare: Methods, Challenges, and Applications**

Haoyang Li[1], Jie Xu[2], Kyra Gan[3], Fei Wang[1], Chengxi Zang[1]*

[1]Department of Population Health Sciences, Weill Cornell Medicine, New York, NY, USA

[2]Department of Health Outcomes & Biomedical Informatics, University of Florida, Gainesville, Florida, USA

[3]Cornell Tech, Cornell University, New York, NY, USA

*Corresponding author

E-mail: chz4001@med.cornell.edu

**Abstract:** Federated causal inference enables multi-site treatment effect estimation without sharing individual-level data, offering a privacy-preserving solution for real-world evidence generation. However, data heterogeneity across sites, manifested in differences in covariate, treatment, and outcome, poses significant challenges for unbiased and efficient estimation. In this paper, we present a comprehensive review and theoretical analysis of federated causal effect estimation across both binary/continuous and time-to-event outcomes. We classify existing methods into weight-based strategies and optimization-based frameworks and further discuss extensions including personalized models, peer-to-peer communication, and model decomposition. For time-to-event outcomes, we examine federated Cox and Aalen–Johansen models, deriving asymptotic bias and variance under heterogeneity. Our analysis reveals that FedProx-style regularization achieves near-optimal bias-variance trade-offs compared to naive averaging and meta-analysis. We review related software tools and conclude by outlining opportunities, challenges, and future directions for scalable, fair, and trustworthy federated causal inference in distributed healthcare systems.

**Introduction**

Causal inference lies at the heart of evidence-based clinical research, aiming to answer questions such as whether a treatment improves survival, reduces complications, or benefits of a target patient population. While randomized trials serve as the gold standard for evidence generation, real-world data (RWD) collected in real-world care, including but not limit to electronic health records (EHRs), administrative claims, and registries are increasingly leveraged to emulate target trials and estimate causal effects in real-world practice. However, as data are intrinsically distributed across institutions due to administrative fragmentation, privacy concerns, and regulatory constraints, traditional centralized causal inference pipelines are becoming infeasible.[1] This has led to the emergence of Federated Causal Inference (FCI): a paradigm that aims to estimate causal effects across multiple decentralized data sources without sharing individual-level data.

Federated learning (FL), originally developed for predictive modeling across distributed devices, has recently gained traction in causal inference.[2–8] In FL, each site performs local computation and exchanges only intermediate statistics or model updates with a central server (or its peers), thereby preserving privacy and institutional autonomy. Extending the FL paradigm to causal inference tasks introduces both opportunities and challenges. On one hand, federated frameworks enable large-scale evidence generation across diverse populations, with potentially increased power and generalizability, while respecting data governance and privacy. On the other hand, causal effect estimation especially under confounding, censoring, or structural heterogeneity poses unique statistical and computational difficulties that distinguish it from standard supervised learning.

One of the key complications in federated causal inference is data heterogeneity.[9] Real-world healthcare data often exhibit systematic differences across sites: population demographics and care practice including treatment protocols, measurement practices, and thus outcome prevalence can all vary. Such heterogeneity violates the identically distributed (i.i.d.) assumption commonly made in classical FL framework, leading to bias and inconsistency in naive estimators. Addressing this issue requires robust estimation strategies that account for site-level distribution shifts while still enabling collaborative learning. In healthcare settings, additional challenges include the fragmentation of electronic health records across institutions, strict data governance policies and varying data completeness and coding standards, all of which complicate federated deployment and inference in healthcare.

In this work, we present a review of federated causal inference method. We organize the discussion around the nature of outcomes—binary or continuous versus time-to-event—and the degree of inter-site variability. For each setting, we survey existing estimation approaches and propose a unified taxonomy that categorizes federated methods into weight-based and optimization-based strategies. Furthermore, we extend the discussion to advanced FL architectures, including FedProx-style regularization, personalized federated learning, peer-to-peer communication, and model decomposition. These frameworks offer varying trade-offs in terms of robustness, scalability, and fairness.

To complement the methodological review, we provide theoretical analyses of asymptotic bias and variance under heterogeneous settings. We derive closed-form characterizations for both the Cox proportional hazards model and the Aalen–Johansen estimator in the presence of baseline hazard and covariate distribution heterogeneity. Our findings highlight the superiority of regularized optimization (e.g., FedProx) in achieving consistency and efficiency over naive averaging or meta-analytic aggregation.

Overall, the contributions of this paper are three folds: (i) to clarify the distinct methodological landscape of federated causal inference; (ii) to provide a unified framework for handling heterogeneity in federated settings with novel theoretical analyses of asymptotic bias and variance under heterogeneous settings; and (iii) to identify key challenges and future directions to advance this nascent but impactful area of research. By bridging developments in federated learning, causal inference, and real-world healthcare data and applications, we hope this work can inform both methodological innovation and translational research.

**Results**

Before presenting the detailed results, we provide a summary of the structure and core contributions of this work. Federated Causal Inference (FCI) refers to estimating treatment effects across multiple data-holding sites without pooling individual-level data, thereby preserving privacy while enabling large-scale evidence generation. A key challenge in FCI is heterogeneity across sites, including differences in patient populations, treatment practices, and outcome-generating mechanisms. We define the homogeneous setting as one where identically distributed (i.i.d.) assumption holds among sites. In contrast, the heterogeneous setting captures realistic multi-site scenarios where i.i.d. assumption dost not hold among sites introducing bias and instability in estimators.[10–12] To address this, we organize our analysis along two outcome types binary or continuous outcomes and time-to-event outcomes and study each under both homogeneous and heterogeneous assumptions. Within each category, we categorize federated estimation strategies into weight-based methods, which reweight local estimators after model training, and optimization-based methods, which directly incorporate heterogeneity-aware adjustments into the training objectives. Beyond methodology, we develop a unified theoretical analysis of federated estimators, deriving closed-form expressions for their asymptotic bias and variance under site-level heterogeneity. These theoretical results show that FedProx regularization achieves lower bias and variance than both naïve averaging and meta-analysis, especially under moderate to strong heterogeneity. We also review an emerging ecosystem of software tools that support federated causal inference, ranging from general-purpose federated learning libraries to domain-specific platforms for biomedical data. Together, these components provide a comprehensive roadmap for designing, analyzing, and deploying federated causal inference frameworks across real-world healthcare systems.

**FCI with Binary/Continue Outcome**

**Homogeneous Settings**

In federated causal inference, homogeneous settings refer to the scenario where the underlying data distributions are identical across all participating sites. Specifically, the covariate distributions, treatment assignment mechanisms, and outcome models are assumed to be the same among sites to allow direct aggregation of local estimators without bias. Under this assumption, common causal estimands such as the Average Treatment Effect (ATE) and Conditional Average Treatment Effect (CATE) can be efficiently estimated by federated methods that aggregate statistics or model parameters from local computations.

*Average Treatment Effect (ATE).* The ATE measures the expected difference in outcomes if all individuals in the population were assigned to treatment versus control. Let $T \in \{0,1\}$ denote the binary treatment assignment and let $Y \in \mathbb{R}$ denote the observed outcome. The counterfactual outcomes are $Y(1)$ and $Y(0)$, and the ATE is defined as:

$$\text{ATE} = \mathbb{E}[Y(1) - Y(0)] = \mathbb{E}[Y \mid T = 1] - \mathbb{E}[Y \mid T = 0]. \quad (1)$$

Each site $k \in \{1, \dots, K\}$ estimates the ATE using its local data. A common approach is linear regression adjustment:

$$Y_i = \alpha + \beta T_i + \varepsilon_i, \quad (2)$$

where $\beta$ represents the local estimate of the ATE at site $k$, denoted $\hat{\beta}_k$. The global ATE can then be aggregated as a weighted average of local estimates:

$$\hat{\beta}_{\text{global}} = \frac{\sum_{k=1}^{K} n_k \hat{\beta}_k}{\sum_{k=1}^{K} n_k}, \quad (3)$$

where $n_k$ is the sample size at site $k$. This estimator is unbiased under homogeneous conditions and is equivalent to the pooled estimator.

*Conditional Average Treatment Effect (CATE).* The CATE measures the treatment effect conditional on covariates $X$:

$$\text{CATE}(X) = \mathbb{E}[Y(1) - Y(0) \mid X]. \quad (4)$$

In federated settings, CATE[13] is typically estimated using regression models that condition on covariates. Each site fits a local model with parameter $\theta_k$ of the form:

$$Y_i = f_k(X_i, T_i), \quad (5)$$

where $f_k$ may be a parametric model such as a linear regression, or a flexible model such as a neural network. After training locally, the site-specific parameters $\hat{\beta}_k$ are transmitted to the central server. The global CATE model is obtained by aggregating these parameters:

$$\hat{\beta}_{\text{global}}(X) = \frac{\sum_{k=1}^{K} n_k \hat{\beta}_k(X)}{\sum_{k=1}^{K} n_k}. \quad (6)$$

Such an approach works well under homogeneous conditions where the conditional distributions $P(Y \mid X, T)$ are aligned across sites. However, model averaging may fail under strong site-specific variations, motivating more robust strategies in heterogeneous settings.[14–17]

The final CATE estimate is then computed as $\text{CATE}(X) = Y_1 - Y_0$ by Eq. (5).

**Heterogeneous Settings**

In federated causal inference with heterogeneous data sources, the average treatment effect (ATE) estimation poses unique challenges due to the non-identically distributed data across participating sites. Variations in population demographics, treatment assignment mechanisms, outcome prevalence, or measurement practices can introduce bias and instability into naively aggregated estimators. Existing approaches to tackle heterogeneity in federated ATE estimation can be broadly categorized into **weight-based methods** and **optimization-based methods**. Specifically, weight-based methods adjust the contribution of each site after local estimation, often relying on how similar the site's data distribution is to a reference population. These methods are generally simple and communication-efficient but may be sensitive to noise or outliers. In contrast, optimization-based methods modify the learning objective itself to account for heterogeneity during model training, which can lead to more stable and robust estimates, though they often require more computation and iterative coordination across sites.

**Weight-Based Methods.** Weight-based methods address heterogeneity by adjusting the influence of each site's estimator based on its similarity to a global distribution or reliability of local estimates. These methods operate in a post-estimation regime and focus on re-weighting site-level ATE estimates to form a more accurate global estimate.[18]

*Kernel-Based Weighting.* This approach uses a kernel function to assess the similarity between a site's covariate distribution and a global reference distribution.[19] The intuition is that sites whose data distributions are more aligned with the global distribution should receive higher weights in the final aggregation.[20]

Let $X_k$ denote the empirical distribution of covariates at site $k$, and $X_{global}$ the pooled or reference distribution approximating the pooled population. The weight $W_k$ for site $k$ is defined as:

$$W_k = \exp\left(-\frac{\|X_k - X_{\text{global}}\|^2}{2\sigma^2}\right), \quad (7)$$

where $\sigma$ is a bandwidth parameter controlling the smoothness of the weighting function. The final ATE is computed as a weighted average:

$$\widehat{\text{ATE}}_{extglobal} = \frac{\sum_{k=1}^{K} W_k \cdot \widehat{\text{ATE}}_k}{\sum_{k=1}^{K} W_k}. \quad (8)$$

*Distance-Based Weighting.* Instead of kernels, this method directly computes distances between local and global distributions using metrics such as Mahalanobis or Euclidean distance:

$$W_k = \frac{1}{d(X_k, X_{\text{global}})}, \quad (9)$$

where $d(\cdot,\cdot)$ measures the dissimilarity between distributions. The global ATE is then aggregated as:

$$\widehat{\text{ATE}}_{\text{global}} = \frac{\sum_{k=1}^{K} W_k \cdot \widehat{\text{ATE}}_k}{\sum_{k=1}^{K} W_k}. \quad (10)$$

These methods are lightweight and communication-efficient, making them suitable for large-scale federated networks. However, they rely heavily on accurate estimation of distribution distances, and may be sensitive to high-dimensional noise or outlier sites.

**Optimization-Based Methods.** Optimization-based approaches integrate heterogeneity handling directly into the learning algorithm. A prominent example is the **FedProx** method,[9] which modifies the site-specific objective function by penalizing divergence from the global model.

*FedProx Regularization.* FedProx introduces a proximal term to constrain local model updates:

$$\mathcal{L}_k(\beta) = \mathcal{L}_k^{\text{local}}(\beta) + \frac{\mu}{2}\|\beta - \beta_{\text{global}}\|^2, \quad (11)$$

where $\mathcal{L}_k$ is the loss function of site $k$, $\beta$ is the estimated parameter, and $\mu$ controls the strength of regularization. The global update is performed iteratively as:

$$\beta^{(t+1)} = \beta^{(t)} - \eta \cdot \frac{1}{K}\sum_{k=1}^{K}\left[\nabla \mathcal{L}_k(\beta^{(t)}) + \mu(\beta^{(t)} - \beta_{\text{global}})\right]. \quad (12)$$

This formulation discourages local models from straying too far from a shared reference, thereby stabilizing training across heterogeneous datasets. FedProx is especially beneficial in non-i.i.d. settings or when computational budgets limit multiple communication rounds.

**Table 1.** Summary of weight-based and optimization-based methods

| Category | Method | Handles Heterogeneity | Communication Efficient | Requires Distribution Matching | Robustness to Outliers |
|---|---|---|---|---|---|
| **Weight-based** | Kernel-based weighting | ✓ (via similarity) | ✓ | ✓ | ✗ |
| | Distance-based weighting | ✓ (via dissimilarity) | ✓ | ✓ | ✗ |
| **Optimization-based** | FedProx | ✓ (via regularization) | ✗ (requires iteration) | ✗ | ✓ |

In summary, weight-based methods offer simplicity and communication efficiency but are prone to instability under severe heterogeneity. Optimization-based methods like FedProx offer stronger robustness but may be computationally and communicationally intensive. A promising future direction lies in hybrid schemes that combine the adaptivity of weighted aggregation with the stability of regularized optimization.

## FCI with Time-to-event Outcome

Time-to-event outcomes, or survival outcomes, are common in clinical studies where the timing of an event (e.g., disease progression, death, or recovery) is of primary interest. Federated estimation of causal effects with time-to-event outcomes introduces unique challenges beyond binary or continuous endpoints, particularly in the presence of censored data, site-specific baseline hazards, and competing risks. Two predominant modeling frameworks in this domain are the Cox proportional hazards model[21,22] and the Aalen–Johansen estimator.[23] We review both under homogeneous and heterogeneous assumptions.

### Homogeneous Setting

In the homogeneous setting, the key assumption is that the distribution of covariates, treatment effects, censoring mechanisms, and baseline hazards are consistent across sites. This enables consistent aggregation of local estimators using classical federated learning techniques.

**Cox Proportional Hazards (CoxPH) Model.** The CoxPH model is widely used to estimate the effect of treatment on the hazard rate of an event. At each site $k$, the hazard function for an individual with covariates $X_{ki}$ and treatment $T_{ki} \in \{0,1\}$ is modeled as:

$$h_k(t \mid X_{ki}) = h_0(t)\exp(X_{ki}^\top \beta), \quad (13)$$

where $h_0(t)$ is a shared baseline hazard function across all sites and $\beta$ represents the log hazard ratio. The assumption of a common $h_0(t)$ is valid under the homogeneous setting.

Each site fits the partial likelihood function locally:

$$\mathcal{L}_k(\beta) = \prod_{i=1}^{n_k} \left( \frac{\exp(X_{ki}^\top \beta)}{\sum_{j \in \mathcal{R}_{ki}} \exp(X_{kj}^\top \beta)} \right)^{\delta_{ki}}, \quad (14)$$

where $\delta_{ki}$ is the event indicator and $\mathcal{R}_{ki}$ is the risk set at time $t_{ki}$. The local estimator $\hat{\beta}_k$ can then be transmitted to a central server. If homogeneity holds, i.e., no covariate distribution shifts among sites, the global estimator $\hat{\beta}_{\text{global}}$ can be computed via weighted averaging using the inverse of the local variance estimates or sample sizes:

$$\hat{\beta}_{\text{global}} = \left(\sum_{k=1}^{K} H_k\right)^{-1} \sum_{k=1}^{K} H_k \hat{\beta}_k, \qquad (15)$$

where $H_k$ is the site-specific observed information matrix.[24]

This approach achieves high statistical efficiency and low bias, assuming consistent data generation across sites. The variance of the global estimator is minimized when all sites contribute proportionally, making it analogous to pooled estimation.

**Aalen–Johansen Estimator for Competing Risks.** When multiple mutually exclusive events can occur (e.g., death from different causes), the Aalen–Johansen estimator is used to estimate the cause-specific cumulative incidence function (CIF). In the homogeneous setting, the CIF for event type $j$ at time $t$ is defined as:

$$F_j(t) = \mathbb{P}(T \leq t, \delta = j), \qquad (16)$$

and estimated at each site $k$ by:

$$\hat{F}_{j,k}(t) = \int_0^t \hat{S}_k(u-)d\hat{A}_{j,k}(u), \qquad (17)$$

where $\hat{S}_k(u)$ is the Kaplan–Meier estimate[25] of overall survival, and $\hat{A}_{j,k}(u)$ is the Nelson–Aalen estimate of the cumulative cause-specific hazard for event $j$. Under the homogeneous assumption, these local CIFs can be averaged across sites using sample-size weights or inverse-variance weights to produce a global CIF estimator.

This federated Aalen–Johansen aggregation retains consistency and unbiasedness under the assumption that the true CIF is shared across all centers.

**Heterogeneous Setting**

The heterogeneous setting reflects the reality of most real-world multi-center studies, where differences exist in baseline hazard functions, treatment practices, censoring distributions, and covariate profiles. This makes naive aggregation inappropriate, as direct averaging of estimators leads to biased and inefficient global estimates.

**CoxPH Model under Heterogeneity.** Under heterogeneity, each site $k$ is modeled as:

$$h_k(t \mid X_{ki}) = h_{0k}(t)\exp(X_{ki}^\top \beta), \qquad (18)$$

where $h_{0k}(t)$ varies across sites. While the coefficient vector $\beta$ is assumed to be shared, the baseline hazard heterogeneity leads to different partial likelihoods across sites. Moreover, variation in covariate distributions and treatment assignment mechanisms implies that even the same $\beta$ may yield different interpretations across sites.

Several federated strategies have been proposed to handle this:

- **FedAvg** aggregates local estimates via sample-size weighting:

$$\hat{\beta}_{\text{FedAvg}} = \sum_{k=1}^{K} \frac{n_k}{n} \hat{\beta}_k, \quad (19)$$

which introduces bias if $\mathbb{E}[\hat{\beta}_k] \neq \beta$ due to local estimation inconsistency.

- **FedProx-style Regularization** introduces a proximal penalty in local optimization to enforce coherence with a central model:

$$\mathcal{L}_k^{\text{prox}}(\beta) = \mathcal{L}_k(\beta) + \frac{\mu}{2} \| \beta - \beta_{\text{global}} \|^2, \quad (20)$$

with updates iteratively aggregated to reduce divergence and stabilize estimation.

- **Meta-analysis-based methods** apply inverse-variance or random-effects weighting to mitigate bias from site-specific shifts:

$$\hat{\beta}_{\text{meta}} = \left( \sum_k w_k \right)^{-1} \sum_k w_k \hat{\beta}_k, \quad w_k = \frac{1}{\text{Var}(\hat{\beta}_k) + \tau^2}, \quad (21)$$

with $\tau^2$ capturing between-site heterogeneity.

In recent work, theoretical comparisons show that FedProx-style aggregation outperforms naive FedAvg and meta-analysis in reducing bias and variance under moderate heterogeneity, especially when local estimators are regularized with accurate information weights.

**Aalen–Johansen Estimator under Heterogeneity.** Competing risks estimation under heterogeneity is further complicated by differing event types, incidence patterns, and follow-up durations across sites. Let $F_{j,k}(t)$ denote the true CIF at site $k$, and $b_k(t) = \mathbb{E}[\hat{F}_{j,k}(t)] - F_{j,0}(t)$ be the bias relative to a global target $F_{j,0}(t)$. Naive averaging yields:

$$\hat{F}_j^{\text{FedAvg}}(t) = \sum_{k=1}^{K} \frac{n_k}{n} \hat{F}_{j,k}(t), \quad (22)$$

but this estimator inherits bias if CIFs vary across sites.

Federated alternatives include:

- **Weighted CIF aggregation** based on inverse-variance weights[26]:

$$\hat{F}_j^{\text{meta}}(t) = \sum_{k=1}^{K} \widetilde{w}_k(t) \hat{F}_{j,k}(t), \quad \widetilde{w}_k(t) = \frac{1/V_k(t)}{\sum_{j=1}^{K} 1/V_j(t)}, \quad (23)$$

where $V_k(t)$ is the asymptotic variance of $\hat{F}_{j,k}(t)$.

- **FedProx-like stabilization** based on optimizing local Brier scores or pseudo-residuals while constraining deviation from a global estimator.

Despite these efforts, federated estimation of cause-specific effects under heterogeneous survival remains an open area of research.

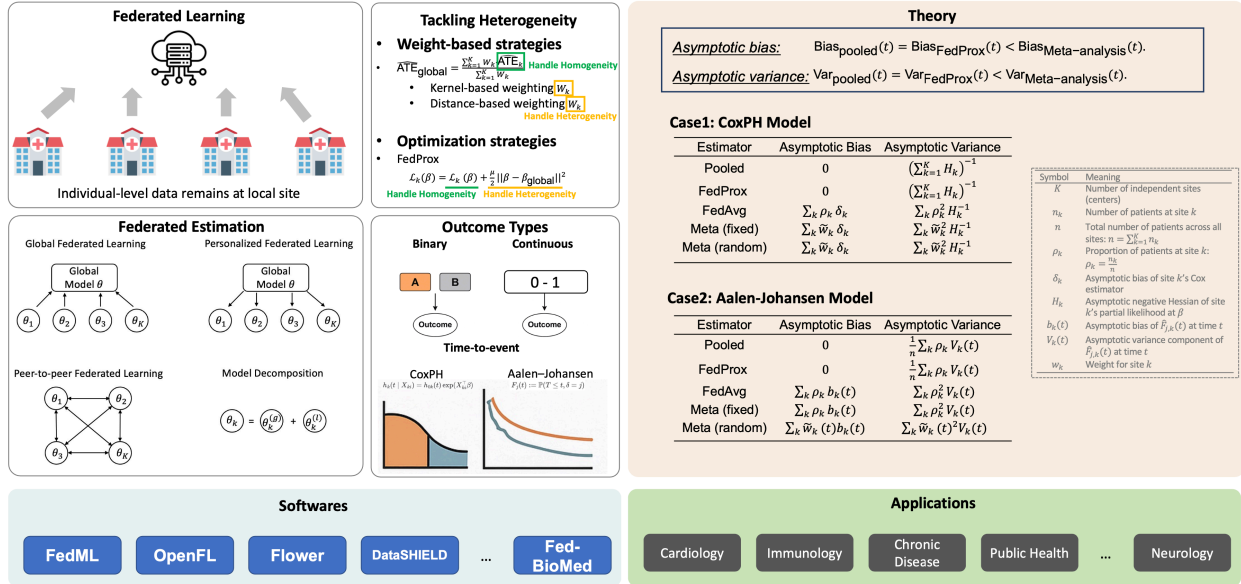

**Figure 1.** Federated causal inference methods framework.

## Federated Estimation framework

In federated causal inference, the design of estimation frameworks plays a critical role in addressing statistical, computational, and communication challenges—especially under heterogeneity. Beyond simple averaging-based approaches, several classes of federated learning (FL) algorithms have emerged, offering various strategies for aggregating models, adapting to local data distributions, and preserving privacy.[27] We summarize four main paradigms: FedProx-based global FL, personalized federated learning, peer-to-peer networks, and model decomposition frameworks.

FedProx is a proximal-gradient-based approach designed to stabilize convergence under data heterogeneity. In this framework, a global model is iteratively optimized by aggregating locally regularized updates, where each site's objective is augmented with a proximity term that penalizes deviation from the global model. This formulation is particularly suited for causal effect estimation in real-world multi-site settings, where treatment assignment and outcome mechanisms differ across locations. By enforcing soft coherence across sites, FedProx offers improved convergence stability and asymptotic efficiency, especially for estimators with identifiable global targets (e.g., average treatment effect, hazard ratios). Moreover, the approach is modular: it applies equally to binary, continuous, or survival outcomes, and accommodates a wide range of estimators including outcome regression, inverse probability weighting (IPW), or doubly robust estimators.

**Algorithm 1: Federated Causal Effect Estimation under Heterogeneity**

**Input**:

- A federated network of $K$ sites, each with local dataset $\mathcal{D}_k$
- Total number of global iterations $T$

**Output**:

- Estimated global parameter $\widehat{\Theta}^{(T)}$, consisting of causal inference model parameters (e.g., $\hat{\theta}^{(T)}$)

---

**Procedure**:

1. Initialize global parameter $\widehat{\Theta}^{(0)} = \theta^{(0)}$
2. **For** $t = 0, 1, \ldots, T-1$ **do**:

    a. Server broadcasts $\theta^{(t)}$ to all $K$ sites

    b. **For each** site $k \in \{1, \ldots, K\}$ **in parallel**:

    i. Each site solves the following proximal local objective:

    $$\theta_k^{(t+1)} = \arg\min_{\theta} \left\{ \mathcal{L}_k(\theta) + \frac{\lambda}{2} \| \theta - \theta^{(t)} \|^2 \right\} \quad (24)$$

    ii. Site $k$ sends updated $\theta_k^{(t+1)}$ to the central server

    c. Server aggregates local updates to obtain the global update:

    $$\theta^{(t+1)} = \sum_{k=1}^{K} \frac{n_k}{n} \theta_k^{(t+1)} \quad (25)$$

    d. Update global parameter: $\widehat{\Theta}^{(t+1)} \leftarrow \theta^{(t+1)}$

3. **Return** optimized global parameter $\widehat{\Theta}^{(T)}$

---

**Notes**:

- $\mathcal{L}_k(\theta)$ denotes the local loss function associated with the causal inference task at site $k$, which may represent an outcome model (e.g., logistic, linear, Cox), a treatment model (e.g., propensity score), or a joint objective.
- $\lambda > 0$ is the FedProx regularization coefficient controlling proximity to the global model.

- This formulation accommodates both binary/continuous and time-to-event outcomes, and can be extended to composite parameter vectors $\theta = (\theta_1, \theta_2)$ (e.g., for jointly learning treatment and outcome models).

This algorithm is applicable across a range of heterogeneous federated causal inference settings, including but not limited to IPW, AIPW, outcome regression, survival models, and time-varying exposure settings.

Personalized FL frameworks aim to explicitly tailor models to individual sites, recognizing that causal relationships may vary locally. Instead of learning a single global model, this paradigm treats the global model as a base initialization and allows each site to fine-tune a local model using local data. The personalization step may involve multiple rounds of local adaptation (e.g., gradient steps) to minimize the site's own loss. This strategy is particularly useful when the conditional average treatment effect (CATE) varies across subpopulations or institutions, and when fairness across sites is prioritized. It enables site-specific inference while still benefiting from shared representation learned across the federation.

**Algorithm 2: Personalized Federated Learning**

**Input:** Same as Algorithm 1, with additional personalization step.

**Initialize:** Global model $\theta^{(0)}$

**Output:** Personalized models $\{\theta_k^{(T)}\}_{k=1}^{K}$

---

**For** $t = 0, 1, \ldots, T-1$ **do**

1. **Server broadcasts** global model $\theta^{(t)}$.

2. **Each site** $k$ initializes $\theta_k^{(t,0)} = \theta^{(t)}$

3. **Local adaptation:** Each site performs multiple gradient steps:

$$\theta_k^{(t,m+1)} = \theta_k^{(t,m)} - \eta \nabla \mathcal{L}_k(\theta_k^{(t,m)}) \quad (26)$$

for $m = 0, \ldots, M-1$, yielding personalized model $\theta_k^{(t+1)}$.

4. **Server aggregates** base parameters to update global model:

$$\theta^{(t+1)} = \sum_{k=1}^{K} \frac{n_k}{n} \theta_k^{(t+1)} \quad (27)$$

**End For**

**Return** final $\theta_k^{(T)}$ at each site.

---

Peer-to-peer (P2P) federated learning removes the central coordinator and adopts a decentralized architecture. Each site iteratively communicates with a set of neighbors, exchanging local models and aggregating them according to predefined weights (e.g., via gossip protocols or consensus algorithms). This design aligns well with scenarios lacking a central server or where network topology is constrained. In the context of causal effect estimation, P2P FL is advantageous in distributed health systems or global networks, where trust, autonomy, and regulatory boundaries inhibit centralized orchestration. It also enables more resilient and fault-tolerant learning by eliminating a single point of failure.

**Algorithm 3: Peer-to-Peer Federated Averaging**

**Input:** Same as Algorithm 1, with decentralized network topology.

**Initialize:** Each site $k$ initializes local model $\theta_k^{(0)}$

**Output:** Locally converged models $\{\theta_k^{(T)}\}_{k=1}^K$

---

**For** $t = 0,1,\ldots,T-1$ **do**

1. **Each site** $k$ updates locally:

$$\theta_k^{(t+1/2)} = \theta_k^{(t)} - \eta \nabla \mathcal{L}_k(\theta_k^{(t)}) \quad (28)$$

2. **Each site exchanges** parameters with its neighbors $\mathcal{N}_k$ and updates by:

$$\theta_k^{(t+1)} = \sum_{j \in \mathcal{N}_k \cup \{k\}} w_{kj}\, \theta_j^{(t+1/2)} \quad (29)$$

where $w_{kj}$ are communication weights.

**End For**

**Return** $\{\theta_k^{(T)}\}$

---

Model decomposition separates the global model into shared and local components, typically through a common representation function (e.g., a feature encoder) and site-specific heads (e.g., prediction layers). Each site optimizes both parts jointly, with the shared component constrained to remain close to the aggregated global representation. This setup allows sites to retain flexibility in their outcome generation mechanisms while sharing knowledge at a representation level. This strategy is particularly attractive when federated tasks span heterogeneous populations or when modeling complex causal structures involving high-dimensional covariates. For instance, a shared encoder can extract common latent confounders, while site-specific heads account for differences in treatment protocols or population-level effect modifiers.

**Algorithm 4: Federated Model Decomposition**

**Input:** Global representation model $f$, site-specific heads $\{h_k\}$, number of rounds $T$

**Initialize:** Shared model $f^{(0)}$, local heads $h_k^{(0)}$

**Output:** Final shared model $f^{(T)}$, personalized heads $\{h_k^{(T)}\}$

---

**For** $t = 0, 1, \ldots, T-1$ **do**

1. **Server sends** current shared model $f^{(t)}$ to all sites.
2. **Each site** updates head and shared model jointly by solving:

$$(f_k^{(t+1)}, h_k^{(t+1)}) = \arg\min_{f,h} \mathcal{L}_k(f, h) + \frac{\lambda}{2} \| f - f^{(t)} \|^2 \quad (30)$$

3. **Server aggregates** shared model:

$$f^{(t+1)} = \sum_{k=1}^{K} \frac{n_k}{n} f_k^{(t+1)} \quad (31)$$

**End For**

**Return** $f^{(T)}, \{h_k^{(T)}\}$

---

## Theoretical Foundations of Federated Estimators for Binary/Continuous Outcomes

We summarize the conclusions on the asymptotic properties of federated ATE estimators from Khellaf et al.[28] as follows.

We consider a set of $K$ independent studies. For each individual $i$, the observed data is: $Z_i = (X_i, W_i, Y_i, H_i)$ where $X_i \in \mathbb{R}^d$ is the covariate vector, $W_i \in \{0,1\}$ is the treatment assignment, $Y_i \in \mathbb{R}$ is the observed outcome, $H_i \in \{1, \ldots, K\}$ is the study index.

Our goal is to estimate the **Average Treatment Effect (ATE)** across the population $\tau = \mathbb{E}[\mathbb{E}[Y(1) - Y(0) \mid H]]$

We assume a **linear outcome model** for each treatment arm $w \in \{0,1\}$: $Y_{k,i}(w) = c^{(w)} + X_{k,i}^\top \beta^{(w)} + \epsilon_{k,i}^{(w)}$ where $c^{(w)} \in \mathbb{R}$ is the intercept, $\beta^{(w)} \in \mathbb{R}^d$ are the coefficients, $\epsilon_{k,i}^{(w)}$ are zero-mean noise terms with variance $\sigma^2$.

The ATE can be rewritten compactly as: $\tau = \mathbb{E}[X_{i'}](\theta^{(1)} - \theta^{(0)})$, where $\theta^{(w)} = (c^{(w)}, \beta^{(w)})$ and $X_{i'} = (1, X_i)$

We first make the summary of the key notations and provide two assumptions.

| Symbol | Meaning |
|---|---|
| $K$ | Number of studies |
| $n$ | Total number of observations |
| $n_k$ | Number of observations in study $k$ |
| $n_k^{(w)}$ | Number of observations in study $k$ assigned to treatment $w$ |
| $\rho_k$ | Probability that a sample comes from study $k$, i.e., $\rho_k = \frac{n_k}{n}$ |
| $p_k$ | Treatment probability in study $k$ |
| $p$ | Overall pooled treatment probability: $p = \sum_k \rho_k p_k$ |
| $X_{k,i}$ | Covariates of individual $i$ in study $k$ |
| $\beta^{(w)}$ | True coefficient vector for treatment arm $w$ |
| $\theta^{(w)} = (c^{(w)}, \beta^{(w)})$ | Augmented coefficient vector including intercept |
| $\Sigma$ | Covariance matrix of $X$, i.e., $\Sigma = \mathrm{Var}(X)$ |
| $\|\cdot\|_\Sigma^2$ | Mahalanobis norm: $\| v \|_\Sigma^2 = v^\top \Sigma v$ |

**Assumptions**

**Condition 1 (Local Full Rank)**

Each local study satisfies: $\mathrm{rank}\left(X_k^{(w)\top} X_k^{(w)}\right) = d, \quad \forall k, w$, i.e., $n_k^{(w)} \geq d$.

**Meaning**: Each study can independently estimate a full-rank OLS model.

**Condition 2 (Federated Full Rank)**

At the pooled level: $\mathrm{rank}\left(X^{(w)\top} X^{(w)}\right) = d, \quad \forall w$ i.e., $\sum_k n_k^{(w)} \geq d$.

**Meaning**: Only the **combined** data is sufficient for OLS estimation, local studies may be underpowered.

Considering **Local Estimator**

Each study $k$ fits OLS models and estimates:

$$\hat{\tau}_k = \frac{1}{n_k} \sum_{i=1}^{n_k} \left( X_{k,i'}^\top \hat{\theta}_k^{(1)} - X_{k,i'}^\top \hat{\theta}_k^{(0)} \right) \quad (1)$$

**Intuition**: Purely local information, no collaboration. High variance due to small $n_k$.

**Meta-Analysis Estimators**

Aggregate local $\hat{\tau}_k$ at a central server.

- **Meta-SW** (sample-size weighted): $\hat{\tau}_{\text{Meta-SW}} = \sum_{k=1}^{K} \frac{n_k}{n} \hat{\tau}_k$

- **Meta-IVW** (inverse-variance weighted): $\hat{\tau}_{\text{Meta-IVW}} = \frac{\sum_k V(\hat{\tau}_k)^{-1} \hat{\tau}_k}{\sum_k V(\hat{\tau}_k)^{-1}}$

**Intuition**:

- **Meta-SW** is simple and always unbiased.

- **Meta-IVW** minimizes variance among aggregation schemes but needs estimated variances and can be biased if covariate shifts exist.

**One-Shot Federated Estimators**

**Two communication rounds**:

4. Share local model parameters $\hat{\theta}_k^{(w)}$, federate them into $\hat{\theta}_{1S}^{(w)}$.

5. Each study computes ATE using the federated models.

- **1S-SW**: $\hat{\tau}_{\text{1S-SW}} = \sum_{k=1}^{K} \frac{n_k}{n} \hat{\tau}_{\text{1S-SW},k}$

- **1S-IVW**: $\hat{\tau}_{\text{1S-IVW}} = \sum_{k=1}^{K} \frac{n_k}{n} \hat{\tau}_{\text{1S-IVW},k}$

**Intuition**:

- Federate richer **model** information (not just scalar ATEs), improving precision.

- **1S-IVW** achieves **pooled-level efficiency** if Condition 1 holds.

**Gradient-Based Federated Estimator (GD)**

Using **FedAvg** style gradient descent across studies to fit the pooled OLS solution:
$\hat{\tau}_{\text{GD}} = \sum_{k=1}^{K} \frac{n_k}{n} \hat{\tau}_{\text{GD},k}$

**Intuition**:

- Allows learning the **global model** without needing each study to be locally full-rank.

- Handles small $n_k$, but needs more communication ($(T+1)$ rounds).

**Asymptotic Variance Comparison**

| Estimator | Condition | Asymptotic Variance $V^\infty$ | Comm. Rounds | Comm. Cost |
|---|---|---|---|---|
| Local ($\hat{\tau}_k$) | Cond. 1 | $\dfrac{\sigma^2}{n_k p_k(1-p_k)} + \dfrac{1}{n_k} \| \beta^{(1)} - \beta^{(0)} \|_\Sigma^2$ | 0 | 0 |
| Meta-SW | Cond. 1 | $\dfrac{\sigma^2}{n} \sum_k \dfrac{\rho_k}{p_k(1-p_k)} + \dfrac{1}{n} \| \beta^{(1)} - \beta^{(0)} \|_\Sigma^2$ | 1 | $O(1)$ |
| 1S-SW | Cond. 1 | $\dfrac{\sigma^2}{np(1-p)} + \dfrac{1}{n} \| \beta^{(1)} - \beta^{(0)} \|_\Sigma^2$ | 2 | $O(d)$ |
| 1S-IVW | Cond. 1 | same as 1S-SW | 2 | $O(d^2)$ |
| GD | Cond. 2 | same as 1S-IVW | $(T+1)$ | $O(Td)$ |
| Pool | Cond. 2 | same as 1S-IVW | — | — |

**Theoretical analyses**

- **Theorem 1**: One-shot IVW estimator equals pooled estimator under Condition 1.

- **Theorem 2**: Variance ordering:

$$V^\infty(\hat{\tau}_{\text{pool}}) = V^\infty(\hat{\tau}_{\text{GD}}) = V^\infty(\hat{\tau}_{\text{1S-IVW}}) \leq V^\infty(\hat{\tau}_{\text{Meta-IVW}}) \leq V^\infty(\hat{\tau}_{\text{Meta-SW}}) \quad (2)$$

- **Theorem 3** (Under Covariate Shift): Meta-IVW becomes biased. Pool, GD, and 1S-IVW remain unbiased.

Overall, local estimators rely purely on within-study information without any collaboration across sites, resulting in poor statistical efficiency especially when the local sample size is small. Meta-SW is a simple and safe method when studies are heterogeneous because it avoids complex modeling assumptions, but it generally suffers from larger asymptotic variance. Meta-IVW can achieve better efficiency compared to Meta-SW by leveraging inverse variance weighting, but it is sensitive to covariate distribution shifts across studies, leading to potential bias. One-shot IVW federated estimators, if each site has enough sample size to ensure local full-rank conditions, can match the pooled estimator in statistical efficiency with only limited communication, making them highly attractive. Finally, the gradient-based federated estimator (GD) is the most robust approach when local sample sizes are small, achieving pooled-level efficiency at the cost of iterative communication rounds across sites.

## Theoretical Foundations of Federated Estimators for Time-to-Event Outcomes

We consider a federated learning framework involving $K$ independent sites, each collecting time-to-event data from a local patient population. Let site $k \in \{1, \ldots, K\}$ have sample size $n_k$, and let the total number of patients be $n = \sum_{k=1}^{K} n_k$. The data at each site is modeled according to either a Cox proportional hazards model or a competing risks multistate process for nonparametric cumulative incidence estimation. We aim to characterize the asymptotic properties of various federated estimators under these two paradigms.

### Definition 1 (Cox Model)

For each subject $i$ at site $k$, let $T_{ki}$ denote the observed event or censoring time, $\delta_{ki} \in \{0,1\}$ the event indicator, and $X_{ki} \in \mathbb{R}^d$ the covariate vector. We assume the event time distribution at site $k$ follows a Cox proportional hazards model of the form:

$$h_k(t \mid X_{ki}) = h_{0k}(t)\exp(X_{ki}^\top \beta), \quad (32)$$

where $h_{0k}(t)$ is a site-specific, unspecified baseline hazard function, and $\beta \in \mathbb{R}^d$ is a global coefficient vector that is assumed to be common across all centers.

Let $\hat{\beta}_k$ be the maximum partial likelihood estimator at site $k$, and define:

$$\delta_k := \lim_{n_k \to \infty} \mathbb{E}[\hat{\beta}_k] - \beta_0, \quad H_k := \lim_{n_k \to \infty} \left(-\frac{1}{n_k}\nabla^2 \mathcal{L}_k(\beta_0)\right), \quad \rho_k := \frac{n_k}{n}. \quad (33)$$

### Definition 2 (Aalen–Johansen Estimator for Competing Risks)

Let $T_{ki} \in [0, \tau]$ be the observed time, and $\delta_{ki} \in \{0,1,\ldots,J\}$ the event type. The cumulative incidence function (CIF) for event type $j$ is:

$$F_j(t) := \mathbb{P}(T \leq t, \delta = j). \quad (34)$$

The Aalen–Johansen estimator at site $k$ is:

$$\hat{F}_{j,k}(t) = \int_0^t \hat{S}_k(u-)\, d\hat{A}_{j,k}(u), \quad (35)$$

where $\hat{S}_k(u)$ is the Kaplan–Meier estimator, and $\hat{A}_{j,k}(u$ is the Nelson–Aalen estimator for cause $j$. Define:

$$b_k(t) := \mathbb{E}[\hat{F}_{j,k}(t)] - F_{j,0}(t), \quad V_k(t) := \lim_{n_k \to \infty} n_k \cdot \mathrm{Var}(\hat{F}_{j,k}(t)). \quad (36)$$

### Definition 3 (Weighting Schemes)

Let $w_k = \frac{1}{\mathrm{Var}(\hat{\theta}_k)} \approx n_k H_k$. For random-effects meta-analysis:

$$w_k(t) := \left(\frac{V_k(t)}{n_k} + \tau^2\right)^{-1}, \quad \tilde{w}_k(t) := \frac{w_k(t)}{\sum_{j=1}^{K} w_j(t)}. \quad (37)$$

**Definition 4 (Site-Level Heterogeneity)**

We define heterogeneity across sites to refer to systematic differences in the data-generating processes, as shown in Figure 2. Specifically, in the Cox model, heterogeneity may arise from two sources:

- **Baseline hazard heterogeneity**: The baseline hazard function $h_{0k}(t)$ differs across sites, i.e., $h_{0k}(t) \neq h_{0k'}(t)$ for $k \neq k'$.

- **Covariate distribution shifts**: The distribution of $X_{ki}$ varies by site, i.e., $P_k(X) \neq P_{k'}(X)$.

These types of heterogeneity induced by site variable $H$ imply that even under a shared global coefficient $\beta$, the marginal hazard or survival curves may differ between sites. In the Aalen–Johansen setting, heterogeneity refers to variation in the marginal cumulative incidence functions $F_{j,k}(t)$ across sites, due to differences in event rates or censoring mechanisms.

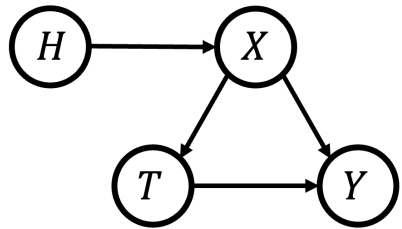

**Figure 2.** Graphical model in the heterogeneous distributions setting.

Such heterogeneity directly impacts the bias and efficiency of federated and meta-analysis estimators, and motivates the theoretical comparison presented above.

**Theorem 1 (Cox Model: Asymptotic Bias and Variance)**

Under the assumptions of the Cox model and assuming heterogeneous baseline hazards, the asymptotic bias and variance of the Cox estimators are summarized as:

**Table 2.** Asymptotic bias and variance of CoxPH model under different methods

| Estimator | Asymptotic Bias | Asymptotic Variance | |
|---|---|---|---|
| Pooled | 0 | $\left(\sum_{k=1}^{K} H_k\right)^{-1}$ | |
| FedProx (Our FL) | 0 | $\left(\sum_{k=1}^{K} H_k\right)^{-1}$ | (38) |
| FedAvg | $\sum_k \rho_k \delta_k$ | $\sum_k \rho_k^2 H_k^{-1}$ | |
| Meta (fixed) | $\sum_k \widetilde{w}_k \delta_k$ | $\sum_k \widetilde{w}_k^2 H_k^{-1}$ | |
| Meta (random) | $\sum_k \widetilde{w}_k \delta_k$ | $\sum_k \widetilde{w}_k^2 H_k^{-1}$ | |

**Theorem 2 (Aalen–Johansen: Asymptotic Bias and Variance at Time $t$)**

Assuming non-informative right censoring and standard convergence conditions for the Aalen–Johansen estimator, the asymptotic properties of cumulative incidence estimation at time $t \in [0, \tau]$ are as follows:

**Table 3.** Asymptotic bias and variance of Aalen–Johansen model under different methods

| Estimator | Asymptotic Bias at $t$ | Asymptotic Variance at $t$ | |
|---|---|---|---|
| Pooled | 0 | $\frac{1}{n}\sum_k \rho_k V_k(t)$ | |
| FedProx (Our FL) | 0 | $\frac{1}{n}\sum_k \rho_k V_k(t)$ | |
| FedAvg | $\sum_k \rho_k b_k(t)$ | $\sum_k \rho_k^2 V_k(t)$ | (39) |
| Meta (fixed) | $\sum_k \rho_k b_k(t)$ | $\sum_k \rho_k^2 V_k(t)$ | |
| Meta (random) | $\sum_k \widetilde{w}_k(t) b_k(t)$ | $\sum_k \widetilde{w}_k(t)^2 V_k(t)$ | |

Overall, not matter for CoxPH model or Aalen–Johansen model, we have the following conclusions under heterogeneous site setting:

$$\text{Bias}_{\text{pooled}}(t) = \text{Bias}_{\text{FedProx}}(t) = 0$$
$$< \text{Bias}_{\text{FedAvg}}(t), \text{Bias}_{\text{Meta (fixed)}}(t), \text{Bias}_{\text{Meta (random)}}(t). \quad (40)$$

$$\text{Var}_{\text{pooled}}(t) = \text{Var}_{\text{FedProx}}(t) < \text{Var}_{\text{FedAvg}}(t), \text{Var}_{\text{Meta (fixed)}}(t), \text{Var}_{\text{Meta (random)}}(t). \quad (41)$$

**Software Tools Supporting Federated Causal Inference**

A growing ecosystem of federated learning (FL) software is emerging to support causal inference across distributed data sources. These tools span general-purpose FL platforms, privacy-enhancing infrastructures, and domain-specific toolkits tailored to medical and survival data. The table below summarizes representative software packages relevant to federated causal inference:

**Table 4.** Federated causal inference related softwares

| Software Name | Key Feature / Area | URL / Reference |
|---|---|---|
| **FedML** | General-purpose FL framework with simulation and deployment support | fedml.ai |
| **OpenFL** | Intel's general FL library with strong orchestration tools | intel/openfl |
| **Flower** | Flexible FL framework supporting cross-platform deployment | flower.dev |
| **DataSHIELD** | Federated statistical analysis in R, widely used in epidemiology | datashield.org |
| **VANTAGE6** | Privacy-preserving framework for health research | vantage6.ai |
| **dsSurvival** | R-based toolkit for federated survival models | dsSurvival |
| **Flamby** | Benchmark datasets for federated healthcare tasks | FLamby |
| **Fed-BioMed** | Biomedical FL platform with PyTorch integration | Fed-BioMed |

These platforms differ in terms of programming language (e.g., Python, R), target application domain (e.g., survival analysis, biomedicine), and technical features (e.g., support for secure aggregation, cross-silo deployment). Among them, **FLAME** and **dsSurvival** provide the most direct support for causal inference tasks, whereas **FedML**, **Flower**, and **OpenFL** offer modular backbones that can be extended to implement custom estimators such as ATE, CATE, or time-to-event models. Tools like **DataSHIELD** and **VANTAGE6** emphasize secure computation and policy compliance, making them particularly suited for multi-site clinical collaborations.

Evaluating the performance of federated causal inference methods requires careful consideration of both causal estimation quality and federated system constraints. Standard metrics from causal inference, including bias, variance, mean squared error (MSE), and coverage probability of confidence intervals, remain central for assessing estimation accuracy. In particular, for binary and continuous outcomes, estimation of average treatment effects (ATE) or conditional average treatment effects (CATE) is typically benchmarked using absolute bias and relative efficiency **compared to pooled analysis.** For time-to-event outcomes, metrics such as hazard ratio bias, concordance index (C-index), and calibration curves are commonly employed. In addition to statistical metrics, federated settings introduce additional dimensions of evaluation. Communication efficiency (e.g., number of communication rounds, transmitted model size), privacy preservation (e.g., differential privacy guarantees, robustness to membership inference attacks), and robustness to heterogeneity (e.g., performance under non-IID covariate or treatment distributions across sites) are critical aspects to benchmark. Benchmarking federated causal inference methods typically involves simulation studies and real-world experiments. Simulation studies enable controlled variation of factors such as sample size, covariate shift, and treatment effect heterogeneity, facilitating systematic comparison across methods. Real-world benchmarking relies on multi-site observational datasets, including federated healthcare repositories or distributed electronic health record (EHR) networks. Open-source resources such as Flamby provide standardized federated datasets (e.g., for cancer survival prediction) that support reproducible evaluation. Researchers often report both pooled ground-truth comparisons (i.e., performance relative to centrally aggregated data) and meta-analysis baselines to demonstrate the advantages of federated learning over traditional aggregation methods.

Thus, a comprehensive evaluation of federated causal inference methods demands a multi-faceted approach, integrating causal accuracy, system efficiency, privacy assurance, and robustness under realistic deployment scenarios.

| | Meta-analysis | Traditional FL | FedProx-based FL |
|---|---|---|---|
| Handling homogeneous site data | 🟢 | 🟢 | 🟢 |
| Unbalanced sample sizes (Fairness across sites) | 🔴 | 🟡 | 🟢 |
| Global confounder adjustment | 🔴 | 🟡 | 🟢 |
| Convergence stability (Less asymptotic bias) | 🔴 | 🔴 | 🟢 |
| Statistical efficiency (Less asymptotic variance) | 🔴 | 🔴 | 🟢 |
| Communication efficiency | 🟢 | 🟡 | 🟡 |
| Privacy mechanisms | 🟢 | 🟡 | 🟢 |
| Robustness to missingness | 🔴 | 🟡 | 🟢 |

**Figure 2.** Characteristics of meta-analysis, traditional federated learning methods (e.g., FedAvg, FedSGD) and FedProx-based federated learning method. Colors indicate whether these methods have this feature (green), partially have this feature (amber) or barely or do not have this feature (red).

**FCI for Healthcare Applications**

Federated causal inference (FCI) methods have gained increasing traction in biomedical research, particularly in scenarios where large-scale treatment effect estimation must be performed across decentralized, privacy-sensitive data sources.[29] By enabling collaboration across institutions without requiring data pooling, FCI offers an effective framework for generating generalizable and regulatory-compliant clinical evidence. Below, we summarize major application domains, focusing on both the data modality (e.g., EHR, survival data) and disease-specific use cases.[30,31]

A prominent application domain is **multi-institutional electronic health records (EHR)**.[32] Large consortia such as OHDSI, PCORnet, and TriNetX have enabled the deployment of federated algorithms across hospital networks, allowing for real-world treatment effect estimation without data centralization. In this context, federated causal models have been used to estimate the average treatment effect (ATE) and conditional average treatment effect (CATE) for interventions ranging from antihypertensive medications to mental health therapies. These studies often leverage federated implementations of outcome regression, inverse probability weighting, and doubly robust estimators, using frameworks such as FedAvg, FedProx, or personalized FL. Such deployments are particularly impactful for rare diseases or low-resource sites, where local sample sizes are insufficient for inference but collaboration through federation can yield statistically valid estimates.

Another major direction is **federated survival analysis for treatment effect estimation**, where the outcome of interest is the time to an event such as death, hospitalization, or relapse. Cox proportional hazards models and Aalen–Johansen estimators have been adapted to federated settings, enabling collaborative modeling of censored data across institutions with differing patient follow-up patterns. These methods are vital in evaluating long-term treatment effectiveness, such as the impact of statins on cardiovascular outcomes or immunotherapies on cancer survival. By integrating site-level heterogeneity into the estimator design—either through regularization (as in FedProx) or decomposition (as in model-split FL)—federated survival methods achieve a balance between privacy, robustness, and interpretability.

FCI has also been applied to **drug repurposing and federated causal discovery**, where the goal is to identify existing compounds that may have unintended therapeutic benefits. Here, FCI frameworks incorporate causal graph discovery algorithms, either constraint-based or score-based, to infer potential treatment-outcome pathways across distributed datasets. For instance, recent efforts in federated Alzheimer's disease research have explored causal links between anti-diabetic medications and cognitive outcomes using privacy-preserving DAG learning, guiding repurposing hypotheses that are subsequently tested in prospective studies. These applications highlight the synergy between causal reasoning and cross-institutional collaboration, especially when experimental studies are infeasible.

From a disease perspective, **FCI has been widely adopted during the COVID-19 pandemic**, where institutions urgently needed to estimate treatment effects (e.g., remdesivir, dexamethasone) across geographies while preserving patient privacy. FCI methods were used to emulate target trials under real-world conditions, accounting for differences in local protocols and data quality. This line of work demonstrates the practicality of federated causal pipelines for outbreak response and global health analytics.[33]

Beyond infectious diseases, FCI is increasingly used in **chronic disease management**—particularly in cardiovascular, renal, and metabolic diseases—where longitudinal EHR data is abundant and site-level practice heterogeneity is substantial. Federated causal methods enable robust treatment effect estimation under such heterogeneity, for example, by modeling center-specific hazards or applying personalized adaptation at each hospital.

In **oncology**, FCI facilitates multi-center collaboration to assess real-world effectiveness of cancer therapies across diverse populations. Applications include evaluating immunotherapy in non-small cell lung cancer, estimating progression-free survival in breast cancer, and comparing treatment lines in colorectal cancer. Survival-based federated methods are especially relevant here, given the time-dependent nature of oncology endpoints and the heterogeneity in biomarker usage, staging criteria, and treatment lines across sites.

Finally, **neurological and mental health disorders** such as Alzheimer's disease,[34–36] depression,[37–39] and schizophrenia[40,41] present a natural application area for FCI due to the sensitivity of the data and the need for cross-site generalizability. Distributed neuroimaging studies and federated EHR-based emulations of psychiatric interventions have demonstrated the feasibility of estimating causal effects while protecting patient confidentiality.

In sum, federated causal inference is emerging as a foundational tool for collaborative clinical research across the healthcare continuum. Its capacity to support large-scale, privacy-preserving, and heterogeneity-aware causal estimation renders it particularly well-suited for regulatory science, real-world evidence generation, and equitable treatment evaluation across diverse clinical settings.

## Opportunities

The advent of federated causal effect estimation frameworks offers unprecedented opportunities for advancing clinical research and real-world evidence generation across distributed, heterogeneous data environments. As multi-center datasets become increasingly prevalent in healthcare, the need for privacy-preserving, bias-resilient, and interpretable causal inference frameworks is growing rapidly. Federated learning (FL), particularly when adapted for causal inference under heterogeneity, presents several promising directions that can reshape methodological paradigms and translational practice.[42,43]

### Enabling Multi-site Evidence Generation at Scale

Federated causal inference methods enable rigorous effect estimation across decentralized datasets without requiring central access to sensitive patient-level information. This capacity is critical in modern healthcare systems where data silos are governed by strict privacy regulations and institutional autonomy. By facilitating cross-site collaboration, federated frameworks support large-scale clinical studies, comparative effectiveness research, and post-marketing surveillance in a manner previously limited by logistical and legal constraints. In particular, hospitals, biobanks, and health information exchanges can collaboratively estimate treatment effects for rare subpopulations, long-term outcomes, or site-specific interventions. The ability to generalize findings beyond a single institution enhances external validity and mitigates biases induced by local practice variation.

### Improving Fairness and Representativeness

Traditional pooled analysis or centralized machine learning models often fail to capture underrepresented patient subgroups or site-specific treatment effects. Federated approaches, especially personalized and model decomposition frameworks, enable each site to retain local modeling capacity while contributing to global learning. This enhances fairness by allowing sites with unique populations—such as pediatric hospitals, rural clinics, or underserved communities—to learn locally optimal policies without conforming to a biased global average. Moreover, conditional estimands such as the Conditional Average Treatment Effect (CATE) can be tailored using federated personalization, improving precision medicine efforts that aim to deliver individualized care. By jointly optimizing global and local models, federated frameworks strike a balance between generalizability and site-specific equity.

**Advancing Causal Inference under Real-world Heterogeneity**

Causal inference in real-world settings faces inherent challenges from confounding, selection bias, and temporal drift.[44,45] Federated algorithms like FedProx introduce regularization principles that help stabilize learning under such variability. Meanwhile, decomposition-based frameworks enable separation of shared causal mechanisms (e.g., treatment-outcome relations) from local biases (e.g., measurement error, institutional protocols), offering new avenues for causal transportability. Furthermore, these methods lend themselves to extensions such as time-varying exposures, competing risks, and dynamic treatment regimes—scenarios in which traditional statistical models often break down. By incorporating flexible architectures (e.g., neural CATE estimators, survival transformers) and modular optimization objectives, federated causal frameworks can serve as a scaffold for robust, extensible causal modeling.

**Enhancing Privacy and Compliance in Sensitive Domains**

The privacy-preserving nature of federated learning aligns naturally with the demands of health data governance, including HIPAA in the U.S., GDPR in Europe, and institutional review board (IRB) frameworks worldwide. Unlike centralized pipelines that risk re-identification through data aggregation, federated causal inference performs computations locally and only transmits aggregate or obfuscated statistics. When combined with secure aggregation, differential privacy, or homomorphic encryption, these methods can support highly sensitive studies (e.g., mental health, infectious disease, substance use) with robust privacy guarantees. This infrastructure also opens possibilities for regulatory-grade evidence generation, where causal inference methods can be integrated into learning health systems and real-world data platforms without violating legal or ethical mandates. For example, federated trial emulation pipelines could be used by government agencies, payers, or pharmaceutical sponsors to assess treatment effectiveness in near real-time across heterogeneous populations.

**Catalyzing New Research Frontiers**

The flexibility of federated frameworks has implications that go beyond estimation. For instance, peer-to-peer architectures can be used to model decentralized causal consensus, relevant to federated clinical decision-making in global health contexts. Model decomposition methods can be adapted to discover latent subtypes, enabling personalized policy learning under unmeasured confounding. Furthermore, theoretical guarantees on asymptotic bias and variance enable principled model selection, auditing, and benchmarking—ensuring that FL is not just a black-box workaround, but a rigorously analyzable method class. These advances may also inspire novel hybrid designs that combine experimental and observational data, such as federated extensions of instrumental variable analysis, target trial emulation, and mediation analysis. By bridging theoretical rigor with practical applicability, federated causal inference represents a foundational shift in how treatment effects are studied, validated, and translated across institutions.

**Challenges**

While federated causal inference offers transformative potential in privacy-preserving effect estimation across distributed healthcare systems, it also introduces a unique constellation of challenges. These challenges span across statistical validity, algorithmic stability, privacy-security trade-offs, and operational feasibility. A clear understanding of these limitations is essential for safe and effective real-world deployment.

**Trade-offs between Personalization and Generalization**

Personalized federated learning offers an intuitive solution to local heterogeneity by allowing each site to fine-tune its own model. However, this comes at the cost of interpretability and generalizability of causal effects. When every site learns a distinct conditional effect estimator, it becomes difficult to synthesize findings into a coherent population-level treatment recommendation. Moreover, in settings such as regulatory trials or comparative effectiveness research, stakeholders often demand globally valid effect estimates rather than individualized models. Balancing personalization with shared representation remains a key open problem. Model decomposition and multi-task learning offer partial remedies but often require careful coordination between shared and local objectives, which is nontrivial in large, asynchronous federated networks.

**Limited Support for Complex Causal Graphs and Mediation Structures**

Most existing federated causal inference frameworks focus on point estimands such as ATE, CATE, or survival curves under unconfoundedness. However, causal questions in healthcare often require more nuanced structures—such as mediation analysis, dynamic treatment regimes, or instrumental variable (IV) estimation. Encoding such causal graphs into federated protocols is challenging due to the distributed nature of variables (e.g., treatment at site A, mediator at site B), partial observability, and inconsistency in measurement across institutions. Furthermore, ensuring valid identifiability conditions—such as sequential ignorability or exclusion restriction—in federated setups requires formal cross-site assumption coordination, which is rarely guaranteed.

**Communication Constraints and System Scalability**

Federated algorithms incur communication overhead from repeated model exchanges and synchronization across sites. While one-shot or few-shot approaches (e.g., summary-statistic-based FL) reduce this burden, they often sacrifice modeling flexibility or convergence stability. High communication cost also impairs scalability, particularly when participating sites have bandwidth limitations, firewalls, or operate under asynchronous availability. Decentralized approaches (e.g., peer-to-peer FL) partially resolve these issues but introduce new complications, such as consensus instability, gradient staleness, and vulnerability to information drift in loopy topologies. These trade-offs complicate protocol design, especially in real-world health systems where compute and bandwidth budgets are limited and heterogeneous.

**Privacy, Security, and Trustworthiness**

Although federated causal inference preserves raw data privacy by default, it is not inherently immune to information leakage. Model updates, summary statistics, and gradients can still leak sensitive information through model inversion attacks, membership inference, or cross-site correlation analysis. Integrating formal privacy-preserving techniques—such as differential privacy (DP), secure multiparty computation (SMPC), or homomorphic encryption (HE)—adds additional layers of complexity and often degrades model accuracy or convergence speed. In high-stakes domains like oncology or mental health, where the sensitivity of causal conclusions can affect treatment policy or insurance coverage, trust in the inference pipeline is critical. Transparency in the training process, reproducibility of estimands, and verifiability of convergence criteria must be maintained—requirements that are nontrivial under federated optimization with adaptive weighting and regularization.

**Evaluation and Benchmarking Gaps**

Unlike standard supervised learning, evaluating causal inference methods—especially in a federated setting—lacks a universally accepted gold standard. Ground-truth treatment effects are rarely available, and counterfactual outcomes cannot be directly observed. Simulation-based validation is common but often fails to capture the complexity of real-world heterogeneity. Moreover, most publicly available benchmark datasets are either single-site or not designed for federated causal inference. The absence of standardized federated datasets with known heterogeneity profiles hinders meaningful benchmarking, ablation studies, and fair comparison of different methods. There is a pressing need for open-access federated testbeds with realistic confounding, censoring, and missingness patterns to advance the field.

**Future Work**

Federated causal inference represents a rapidly evolving frontier at the intersection of causal reasoning, distributed optimization, and privacy-preserving computation. As its methodological foundations mature, several critical avenues remain open for future research and innovation. These directions span theoretical, algorithmic, and translational layers, and are essential for the reliable deployment of federated causal inference across diverse, real-world healthcare systems.

**Toward Unified Estimation Frameworks Across Outcome Types**

Current federated methods are typically specialized for binary, continuous, or time-to-event outcomes in isolation, often requiring separate architectures and objective functions. A promising direction is the development of **unified estimation frameworks** that can accommodate diverse outcome types—such as recurrent events, ordinal outcomes, and multivariate endpoints—within a single federated protocol. For example, extending the FedProx paradigm to simultaneously estimate treatment effects on survival and quality-of-life metrics in a multi-outcome joint model would improve coherence and efficiency. Similarly, developing flexible representation learning frameworks that support both cross-sectional and longitudinal data would enable federated causal reasoning over dynamic health trajectories.

**Federated Causal Discovery and Graph Learning**

Most current frameworks assume that the causal structure—i.e., the causal graph or adjustment set—is known or specified a priori. However, in practice, sites often lack consensus on variable relationships, or the causal structure may be partially latent. **Federated causal discovery**, which aims to infer a shared or site-specific causal DAG from distributed observational data, remains an underexplored yet essential problem. Advances in this area could leverage recent developments in differentiable structure learning, invariant causal prediction, or constraint-based discovery, tailored to federated privacy and communication constraints.[46–48] This would facilitate more robust confounder adjustment and inform proper estimand specification in distributed environments.

**Dynamic and Real-Time Causal Inference**

The growing availability of continuous, high-frequency data streams (e.g., EHRs, wearables) motivates **real-time federated causal inference**, where treatment effects are updated online as new data arrives. This setting poses several challenges, such as handling delayed outcome feedback, modeling time-varying confounding, and updating models without destabilizing convergence. One potential direction is combining federated survival models with online learning and Kalman-style updates to adapt hazard estimates in near-real-time. Another is incorporating change-point detection to automatically identify shifts in data distributions or causal mechanisms. Such capabilities would be transformative for adaptive clinical trial monitoring,[49] outbreak response, or just-in-time personalized medicine.

**Integration of Privacy-Utility Trade-offs**

While privacy-preserving technologies such as differential privacy and secure aggregation have been incorporated into federated classification and regression tasks, their integration into **causal inference objectives** remains largely uncharted. Estimands such as ATE, CATE, or CIFs are inherently sensitive to subtle variations in covariates and treatment patterns, which makes them particularly vulnerable to degradation under formal privacy constraints. Future work should systematically characterize **privacy-utility trade-offs** in federated causal estimation—e.g., how differential privacy affects asymptotic variance or bias, or how to design noise-injection strategies that respect identifiability assumptions. Developing privacy-aware estimators that remain valid under bounded perturbation would be especially critical for applications in mental health, genetics, or rare disease analysis.

**Fairness and Equity-Aware Federated Estimation**

A growing body of literature highlights that traditional causal estimators may reproduce or amplify existing health disparities if not carefully designed. This issue is magnified in federated settings, where participating sites may represent systematically marginalized populations with distinct health behaviors, data quality, or treatment accessibility. Future research must focus on **equity-aware federated causal inference**, which explicitly accounts for group-level performance disparities and distributional fairness. For example, one might incorporate constraints that ensure consistent treatment effect estimation across demographic subgroups or enforce balanced weighting across underrepresented sites. Cross-site fairness auditing, group-specific regularization, or federated counterfactual fairness diagnostics are promising methodological avenues.

**Standardized Benchmarks and Deployment Pipelines**

As the field matures, there is an urgent need for **standardized federated causal inference benchmarks**, which combine real-world clinical relevance with controlled heterogeneity and known ground-truth causal effects. Benchmarks should include binary and time-to-event outcomes, simulate missingness and measurement error, and support plug-and-play integration of different FL paradigms. In parallel, building open-source software libraries and deployment pipelines that **bridge the gap from algorithm to implementation**—e.g., containerized federated trials, modular estimand builders, or visualization dashboards—would significantly lower the barrier for adoption in academic health centers and industry partners alike.

**Regulatory Readiness and Human Interpretability**

Finally, realizing the translational impact of federated causal inference requires close collaboration with regulators, clinicians, and stakeholders. Unlike predictive models, causal estimates are directly tied to **policy decisions**, such as treatment reimbursement or trial eligibility. Ensuring that federated estimates are interpretable, auditable, and aligned with regulatory requirements (e.g., FDA, EMA) is non-negotiable. This calls for a parallel investment in **explainability tools**, transparent convergence diagnostics, and certification protocols that assess stability under site inclusion/exclusion, data drift, or heterogeneity severity. Techniques such as Shapley explanations, influence functions, or local effect decomposition may be adapted to support regulatory-grade justifications of causal claims under federated constraints.